\documentclass[twocolumn, doublespacing]{bmcart}

\usepackage[utf8]{inputenc}
\usepackage{amsmath}
\usepackage{amssymb}
\usepackage{cite}
\usepackage{graphicx}
\usepackage{grffile}
\usepackage[pdftex]{hyperref}
\usepackage{subfig}
\usepackage{tabularx}
\usepackage{multirow}
\usepackage{color}
\usepackage{ulem}

\makeatletter
\long\def\@makecaption#1#2{
  \if@figurestar\hsize=\textwidth\fi%
  \@tempdima\hsize%
  \advance\@tempdima by-\figure@sep%
  \advance\@tempdima by-\figure@sep%
  \hsize\@tempdima%

  \vskip\abovecaptionskip
  \parbox[t]{\hsize}{{\floatcaptionname@size #1}\hskip.5em #2\par}%
  \vskip\belowcaptionskip}
\makeatother

\graphicspath{{figures/}}

\begin{document}

\begin{frontmatter}
\begin{fmbox}
\vspace{-2em}
\hfill\textsf{\small Accepted for publication by Journal of Cardiovascular Magnetic Resonance}
\dochead{Research}


\title{Automated cardiovascular magnetic resonance image analysis with fully convolutional networks}


\author[
   addressref={aff1},                   
   corref={aff1},                       
   email={w.bai@imperial.ac.uk}         
]{\inits{W.B.} \fnm{Wenjia} \snm{Bai}}
\author[
   addressref={aff1},
   email={m.sinclair@imperial.ac.uk}
]{\inits{M.S.} \fnm{Matthew} \snm{Sinclair}}
\author[
   addressref={aff1},
   email={g.tarroni@imperial.ac.uk}
]{\inits{G.T.} \fnm{Giacomo} \snm{Tarroni}}
\author[
   addressref={aff1},
   email={o.oktay13@imperial.ac.uk}
]{\inits{O.O.} \fnm{Ozan} \snm{Oktay}}
\author[
   addressref={aff1},
   email={m.rajchl@imperial.ac.uk}
]{\inits{M.R.} \fnm{Martin} \snm{Rajchl}}
\author[
   addressref={aff1},
   email={g.vaillant@imperial.ac.uk}
]{\inits{G.V.} \fnm{Ghislain} \snm{Vaillant}}
\author[
   addressref={aff2},
   email={a.lee@qmul.ac.uk}
]{\inits{A.M.L.} \fnm{Aaron M.} \snm{Lee}}
\author[
   addressref={aff2},
   email={n.aung@qmul.ac.uk}
]{\inits{N.A.} \fnm{Nay} \snm{Aung}}
\author[
   addressref={aff3},
   email={elena.lukaschuk@cardiov.ox.ac.uk}
]{\inits{E.L.} \fnm{Elena} \snm{Lukaschuk}}
\author[
   addressref={aff2},
   email={m.sanghvi@qmul.ac.uk}
]{\inits{M.M.S.} \fnm{Mihir M.} \snm{Sanghvi}}
\author[
   addressref={aff2},
   email={filip@zemrak.co.uk}
]{\inits{F.Z.} \fnm{Filip} \snm{Zemrak}}
\author[
   addressref={aff2},
   email={k.fung@qmul.ac.uk}
]{\inits{K.F.} \fnm{Kenneth} \snm{Fung}}
\author[
   addressref={aff2},
   email={j.paiva@qmul.ac.uk}
]{\inits{J.M.P.} \fnm{Jose Miguel} \snm{Paiva}}
\author[
   addressref={aff3},
   email={valentina@simula.no}
]{\inits{V.C.} \fnm{Valentina} \snm{Carapella}}
\author[
   addressref={aff3},
   email={dryj@yuhs.ac}
]{\inits{Y.J.K.} \fnm{Young Jin} \snm{Kim}}
\author[
   addressref={aff4},
   email={h.suzuki@imperial.ac.uk}
]{\inits{H.S.} \fnm{Hideaki} \snm{Suzuki}}
\author[
   addressref={aff1},
   email={b.kainz@imperial.ac.uk}
]{\inits{B.K.} \fnm{Bernhard} \snm{Kainz}}
\author[
   addressref={aff4},
   email={p.matthews@imperial.ac.uk}
]{\inits{P.M.M.} \fnm{Paul M.} \snm{Matthews}}
\author[
   addressref={aff2},
   email={s.e.petersen@qmul.ac.uk}
]{\inits{S.E.P.} \fnm{Steffen E.} \snm{Petersen}}
\author[
   addressref={aff3},
   email={stefan.piechnik@cardiov.ox.ac.uk}
]{\inits{S.K.P.} \fnm{Stefan K.} \snm{Piechnik}}
\author[
   addressref={aff3},
   email={stefan.neubauer@cardiov.ox.ac.uk}
]{\inits{S.N.} \fnm{Stefan} \snm{Neubauer}}
\author[
   addressref={aff1},
   email={b.glocker@imperial.ac.uk}
]{\inits{B.G.} \fnm{Ben} \snm{Glocker}}
\author[
   addressref={aff1},
   email={d.rueckert@imperial.ac.uk}
]{\inits{D.R.} \fnm{Daniel} \snm{Rueckert}}


\address[id=aff1]{
  \orgname{Biomedical Image Analysis Group, Department of Computing, Imperial College London}, 
  \city{London},                              
  \cny{UK}                                    
}
\address[id=aff2]{%
  \orgname{NIHR Biomedical Research Centre at Barts, Queen Mary University of London},
  \city{London},
  \cny{UK}
}
\address[id=aff3]{%
  \orgname{Division of Cardiovascular Medicine, Radcliffe Department of Medicine, University of Oxford},
  \city{Oxford},
  \cny{UK}
}
\address[id=aff4]{%
  \orgname{Division of Brain Sciences, Department of Medicine, Imperial College London},
  \city{London},
  \cny{UK}
}

\begin{abstractbox}
\begin{abstract} 
\parttitle{Background}
Cardiovascular magnetic resonance (CMR) imaging is a standard imaging modality for assessing cardiovascular diseases (CVDs), the leading cause of death globally. CMR enables accurate quantification of the cardiac chamber volume, ejection fraction and myocardial mass, providing information for diagnosis and monitoring of CVDs. However, for years, clinicians have been relying on manual approaches for CMR image analysis, which is time consuming and prone to subjective errors. It is a major clinical challenge to automatically derive quantitative and clinically relevant information from CMR images.

\parttitle{Methods}
Deep neural networks have shown a great potential in image pattern recognition and segmentation for a variety of tasks. Here we demonstrate an automated analysis method for CMR images, which is based on a fully convolutional network (FCN). The network is trained and evaluated on a large-scale dataset from the UK Biobank, consisting of 4,875 subjects with 93,500 pixelwise annotated images. The performance of the method has been evaluated using a number of technical metrics, including the Dice metric, mean contour distance and Hausdorff distance, as well as clinically relevant measures, including left ventricle (LV) end-diastolic volume (LVEDV) and end-systolic volume (LVESV), LV mass (LVM); right ventricle (RV) end-diastolic volume (RVEDV) and end-systolic volume (RVESV).

\parttitle{Results}
By combining FCN with a large-scale annotated dataset, the proposed automated method achieves a high performance in segmenting the LV and RV on short-axis CMR images and the left atrium (LA) and right atrium (RA) on long-axis CMR images. On a short-axis image test set of 600 subjects, it achieves an average Dice metric of 0.94 for the LV cavity, 0.88 for the LV myocardium and 0.90 for the RV cavity. The mean absolute difference between automated measurement and manual measurement was 6.1 mL for LVEDV, 5.3 mL for LVESV, 6.9 gram for LVM, 8.5 mL for RVEDV and 7.2 mL for RVESV. On long-axis image test sets, the average Dice metric was 0.93 for the LA cavity (2-chamber view), 0.95 for the LA cavity (4-chamber view) and 0.96 for the RA cavity (4-chamber view). The performance is comparable to human inter-observer variability.

\parttitle{Conclusions}
We show that an automated method achieves a performance on par with human experts in analysing CMR images and deriving clinically relevant measures.
\end{abstract}


\end{abstractbox}

\end{fmbox}

\end{frontmatter}



\section*{Background}
An estimated 17.7 million people died from cardiovascular diseases (CVDs) in 2015, representing 31\% of all global deaths \cite{WHO2017}. More people die annually from CVDs than any other cause. Technological advances in medical imaging have led to a number of options for non-invasive investigation of CVDs, including echocardiography, computed tomography (CT), cardiovascular magnetic resonance (CMR) etc., each having its own advantages and disadvantages. Due to its good image quality, excellent soft tissue contrast and absence of ionising radiation, CMR has established itself as the non-invasive gold standard for assessing cardiac chamber volume and mass for a wide range of CVDs \cite{Ripley2016, Fihn2012, McMurray2012}. To derive quantitative measures such as volume and mass, clinicians have been relying on manual approaches to trace the cardiac chamber contours. It typically takes a trained expert 20 minutes to analyse images of a single subject at two time points of the cardiac cycle, end-diastole (ED) and end-systole (ES). This is time consuming, tedious and prone to subjective errors.

Here we propose a computational method which can automatically analyse images at all time points across the cardiac cycle and derive clinical measures within seconds. The accuracy for clinical measures is comparable to human expert performance. The method would assist clinicians in CMR image analysis and diagnosis with an automated and objective way for deriving clinical measures, therefore reducing cost and improving work efficiency. It would also facilitate large-population imaging studies, such as the UK Biobank study, which aims to conduct imaging scans of vital organs for 100,000 subjects \cite{UKBiobank2017}. An automated method is crucial for analysing such a large amount of images and extracting clinically relevant information for subsequent clinical studies.

Machine learning algorithms, especially deep neural networks, have demonstrated great potential, achieving or surpassing human performance in a number of visual tasks including object recognition in natural images \cite{He2015}, Go game playing \cite{Silver2016}, skin cancer classification \cite{Esteva2017} and ocular image analysis \cite{LongE2017}. Previously, neural networks have been explored for CMR image analysis \cite{Avendi2016, Ngo2017, Tran2017, LiemanSifry2017}. Most of these studies either use relatively shallow network architectures or are limited by the size of the dataset. None of them have performed a comparison between neural networks and human performance on this task. In 2016, Kaggle organised the second Data Science Bowl for left ventricular (LV) volume assessment \cite{Kaggle2016}. Images from 700 subjects were provided with the LV volumes, however, none of the images were annotated. In 2017, MICCAI organised the ACDC challenge \cite{ACDC2017}, where a training set of 100 subjects were provided with manual annotation. Lieman-Sifry et al. curated a data set of 1,143 short-axis image scans \cite{LiemanSifry2017}, where most of the images had LV endocardial and right ventricle (RV) endocardial contours annotated but only 22\% had LV epicardial contours annotated.

\begin{figure*}[h!]
  \centering
  \includegraphics[width=15cm]{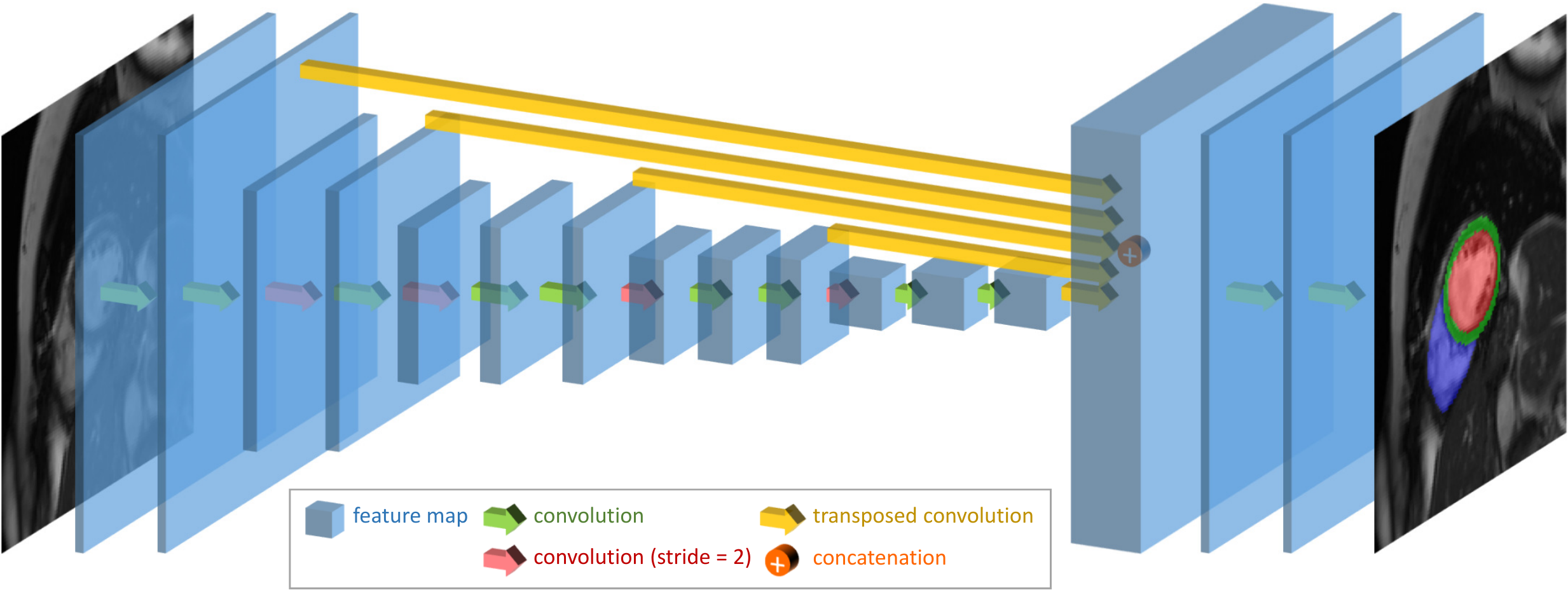}
  \caption{\textbf{The network architecture.} A fully convolutional network is used, which takes the cardiovascular magnetic resonance (CMR) image as input, learns image features from fine to coarse scales through a series of convolutions, concatenates multi-scale features and finally predicts a pixelwise image segmentation. \label{fig:net}}
\end{figure*}

In this paper, we utilise a large dataset of 4,875 subjects with 93,500 images, one or two orders of magnitude larger than previous datasets, and for which all the images have been pixelwise annotated by clinical experts. We trained fully convolutional networks for both short-axis and long-axis CMR image analysis. By combining the power of deep learning and a large annotated dataset for training and evaluation, this paper demonstrated that the proposed automated method can match human-level performance.

\section*{Methods}
\subsection*{Dataset}
The dataset consists of short-axis and long-axis cine CMR images of 5,008 subjects (61.2$\pm$7.2 years, 52.5\% female), acquired from the UK Biobank. The baseline characteristics of the UK Biobank cohort can be viewed in the data showcase at \cite{UKBiobankShowcase2017}. For short-axis images, the in-plane image resolution is 1.8$\times$1.8 mm$^2$ with slice thickness of 8.0 mm and slice gap of 2 mm. A short-axis image stack typically consists of 10 image slices. For long-axis images, the in-plane image resolution is 1.8$\times$1.8 mm$^2$ and only 1 image slice is acquired. Each cardiac cycle consists of 50 time frames. For both short-axis and long-axis views, the balanced steady-state free precession (bSSFP) magnitude images were used for analysis. Details of the image acquisition protocol can be found in \cite{Petersen2016}.

Manual image annotation was undertaken by a team of eight observers under the guidance of three principal investigators and following a standard operating procedure \cite{Petersen2017}. For short-axis images, the LV endocardial and epicardial borders and the RV endocardial borders were manually traced at ED and ES time frames using the cvi$^{42}$ software (version 5.1.1, Circle Cardiovascular Imaging Inc., Calgary, Canada). For long-axis 2-chamber view (2Ch) images, the left atrium (LA) endocardial border was traced. For long-axis 4-chamber view (4Ch) images, the LA and the right atrium (RA) endocardial borders were traced.

In pre-processing, the CMR DICOM images were converted into NIfTI format. The manual annotations from the cvi$^{42}$ software were exported as XML files and also converted into NIfTI format. The images and annotations were quality controlled to ensure that annotations cover both ED and ES frames and without missing slices or missing anatomical structures. For short-axis images, 4,875 subjects (with 93,500 annotated image slices) were available after quality control, which were randomly split into three sets of 3,975/300/600 for training/validation/test, i.e. 3,975 subjects for training the neural network, 300 validation subjects for tuning model parameters, and finally 600 test subjects for evaluating performance. For long-axis 2Ch images, 4,723 subjects were available after quality control, which were split into 3,823/300/600. For long-axis 4Ch images, 4,682 subjects were available, which were split into 3,782/300/600.

\subsection*{Automated image analysis}
For automated CMR image analysis, we utilise a fully convolutional network (FCN) architecture, which is a type of neural network that can predict a pixelwise image segmentation by applying a number of convolutional filters onto an input image \cite{LongJ2015}. The network architecture is illustrated in Figure~\ref{fig:net}. The FCN learns image features from fine to coarse scales using convolutions and combines multi-scale features for predicting the label class at each pixel.

The network is adapted from the VGG-16 network \cite{Simonyan2015} and it consists of a number of convolutional layers for extracting image features. Each convolution uses a 3$\times$3 kernel and it is followed by batch normalisation\footnote{Batch normalisation\cite{Ioffe2015} is a technique which helps address optimisation issues in training deep neural networks, i.e. networks with many layers. It normalises the layer input for each training mini-batch.} and ReLU\footnote{ReLU stands for rectified linear unit. It is a type of activation function for a neuron in artificial neural networks.}. After every two or three convolutions, the feature map is downsampled by a factor of 2 so as to learn features at a more global scale. Feature maps learnt at different scales are upsampled to the original resolution using transposed convolutions\footnote{A transposed convolution is a convolution whose weight matrix has been transposed\cite{TransConv2017}. It is often used for upsampling an image or a feature map.} and the multi-scale feature maps are then concatenated. Finally, three convolutional layers of kernel size 1$\times$1, followed by a softmax function\footnote{Softmax regression is a generalisation of logistic regression to the case where we have multiple classes. It is used for mapping a feature vector to a probability vector.}, are used to predict a probabilistic label map. The segmentation is determined at each pixel by the label class with highest softmax probability. The mean cross entropy between the probabilistic label map and the manually annotated label map is used as the loss function. Excluding the transposed convolutional layers, this network has in total 16 convolutional layers. Details of the network architecture can be found in Table~\ref{tab:net_config}. This architecture is similar to the U-Net\cite{Ronneberger2015}. The main difference is that U-Net performs upsampling step by step. It iteratively upsamples the feature map at each scale by a factor of 2 and concatenates with the feature map at the next scale. In contrast to this, the proposed network may be simpler on the upsampling path. It upsamples the feature map from each scale to the finest resolution in one go and then concatenates all of them.

\begin{table}[h!]
  \caption{\textbf{The network architecture.} The first two columns list the resolution scale and feature map size. The third column lists the convolutional layer parameters, with ``$3 \times 3, 16$" denoting $3 \times 3$ kernel and 16 output features. The last convolutional layer outputs $K$ features, with $K$ denoting the number of label classes.\label{tab:net_config}}
  \renewcommand{\arraystretch}{1.2}
  \centering
  \begin{tabular}{c|c|c}
  \hline 
  scale & size & convolution \\
  \hline 
  1 & 192$\times$192 & $\begin{array}{c}3 \times 3, 16\\ 3 \times 3, 16\end{array}$ \\
  \hline
  2 & 96$\times$96 & $\begin{array}{c}3 \times 3, 32\\ 3 \times 3, 32\end{array}$ \\
  \hline
  3 & 48$\times$48 & $\begin{array}{c}3 \times 3, 64\\ 3 \times 3, 64\\ 3 \times 3, 64\end{array}$ \\
  \hline
  4 & 24$\times$24 & $\begin{array}{c}3 \times 3, 128\\ 3 \times 3, 128\\ 3 \times 3, 128\end{array}$ \\
  \hline
  5 & 12$\times$12 & $\begin{array}{c}3 \times 3, 256\\ 3 \times 3, 256\\ 3 \times 3, 256\end{array}$ \\
  \hline
  \multicolumn{3}{c}{upsample and concatenate} \\
  \multicolumn{3}{c}{scale 1 to 5 features} \\
  \hline
  predict & 192$\times$192 & $\begin{array}{c}1 \times 1, 64\\ 1 \times 1, 64\\ 1 \times 1, K\end{array}$ \\
  \hline
  \end{tabular} 
\end{table}

\subsubsection*{Network training and testing}
Three networks were trained, respectively for segmenting short-axis images, long-axis 2Ch images and 4Ch images. For training each network, all images were cropped to the same size of 192$\times$192 and intensity normalised to the range of $[0,1]$. Data augmentation\footnote{Data augmentation is a technique to increase the size of the training set by applying random spatial transformation or intensity transformation to the original training samples.} was performed on-the-fly, which applied random translation, rotation, scaling and intensity variation to each mini-batch of images before feeding them to the network. Each mini-batch consisted of 20 image slices. The Adam method \cite{Kingma2015} was used for optimising the loss function, with a learning rate of 0.001 and iteration number of 50,000. The method was implemented using Python and TensorFlow. It took about 10 hours to train the VGG-16 network on a Nvidia Tesla K80 GPU.

During the testing stage, it took $\sim$2.2 seconds to analyse the ED and ES time frames of short-axis images for one subject and 9.5 seconds to analyse a full sequence of 50 time frames. For long-axis images, it took $\sim$0.2 seconds to analyse the ED and ES time frames for one subject and 1.4 seconds to analyse a full sequence. It took longer to analyse the short-axis images, because each short-axis image stack typically has 10 slices, whereas a long-axis image stack has only 1 slice.

\begin{figure}[h!]
  \centering
  \includegraphics[width=5cm]{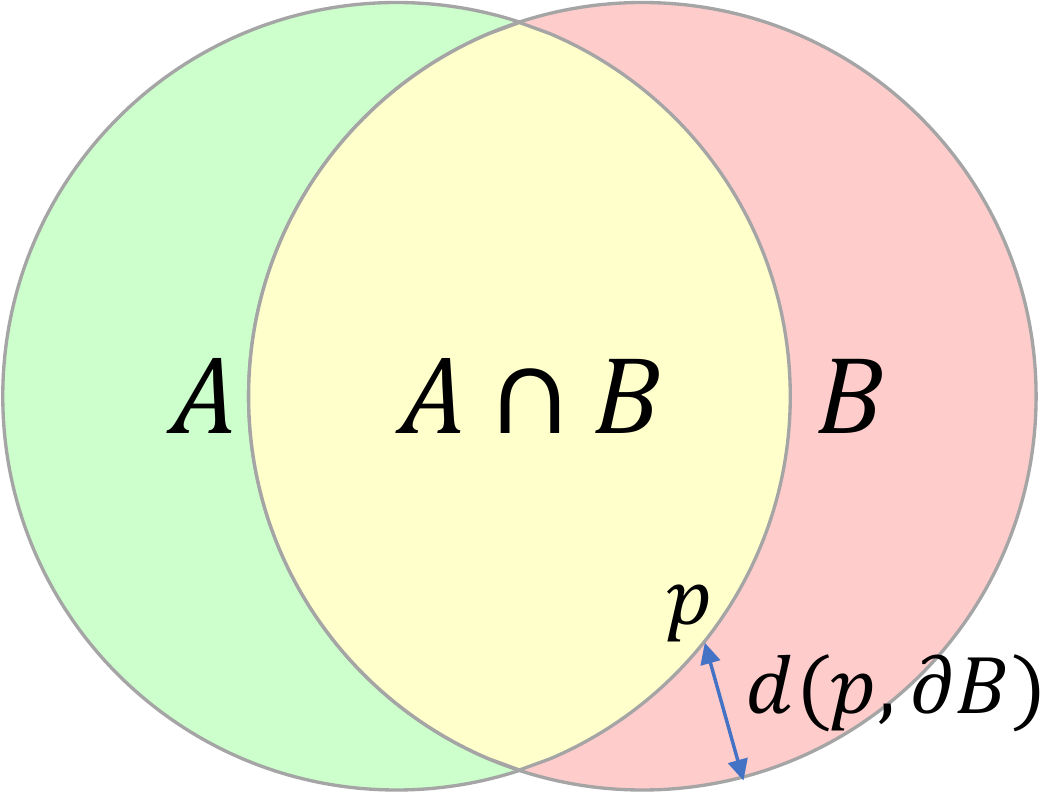}
  \caption{\textbf{Illustration of the Dice metric and contour distance metrics.} $A$ and $B$ are two sets representing automated segmentation and manual segmentation. The Dice metric calculates the ratio of the intersection $|A \cap B|$ over the average area of the two sets $(|A| + |B|)/2$. The mean contour distance first calculates, for each point $p$ on one contour, its distance to the other contour $d(p, \partial)$, then calculates the mean across all the points $p$. The Hausdorff distance calculates the maximum distance between the two contours. \label{fig:dice}}
\end{figure}

\subsubsection*{Evaluation of the method}
For quantitative assessment, we evaluated the performance of the automated method in two ways, respectively using commonly used metrics for segmentation accuracy assessment, including the Dice metric, mean contour distance and Hausdorff distance, and using clinical measures derived from segmentations, including ventricular volume and mass.

Figure~\ref{fig:dice} illustrates the definitions of the Dice metric and contour distance metrics. The Dice metric evaluates the overlap between automated segmentation $A$ and manual segmentation $B$ and it is defined as,
\begin{align}
  \mathrm{Dice} = \frac{2 |A \cap B|}{|A| + |B|} .\nonumber
\end{align}
It is a value between 0 and 1, with 0 denoting no overlap and 1 denoting perfect agreement. The higher the Dice metric, the better the agreement.

The mean contour distance and Hausdorff distance evaluate the mean and the maximum distance respectively between the segmentation contours $\partial A$ and $\partial B$. They are defined as,
\begin{flalign*}
  \mathrm{mean~dist.} & = \frac{1}{2 |\partial A|} \sum_{p \in \partial A} d(p, \partial B)  + \frac{1}{2 |\partial B|} \sum_{q \in \partial B} d(q, \partial A), \nonumber &\\
  \mathrm{Haus.~dist.} & = \max\left(\max_{p \in \partial A} d(p, \partial B), \max_{q \in \partial B} d(q, \partial A)\right), \nonumber
\end{flalign*}
where $d(p, \partial)$ denotes the minimal distance from point $p$ to contour $\partial $. The lower the distance metric, the better the agreement.

We also evaluated the accuracy of clinical measures, which were derived from image segmentations. We calculated the LV end-diastolic volume (LVEDV) and end-systolic volume (LVESV), LV myocardial mass (LVM), RV end-diastolic volume (RVEDV) and end-systolic volume (RVESV) from automated segmentation and compared them to measurements from manual segmentation. The LV and RV volumes were calculated by summing up the number of voxels belonging to the corresponding label class in the segmentation, multiplied by the volume per voxel. The LV mass was calculated by multiplying the LV myocardial volume with the density of 1.05 g/mL \cite{Grothues2002}.

\subsection*{Evaluation of human performance}
For quantitative evaluation of human performance, we assessed the inter-observer variability between manual segmentations by different clinical experts. A set of 50 subjects was randomly selected and each subject was analysed by three expert observers (O1, O2, O3) independently. The Dice metric, contour distance metrics and the difference of clinical measurements were evaluated between each pair of observers (O1 vs O2, O2 vs O3, O3 vs O1).

\subsection*{Qualitative assessment}
As an additional qualitative assessment, two experienced image analysts (respectively with over ten years and four years experiences in cardiovascular image analysis) visually assessed the segmentations for 250 test subjects. According to an in-house standard operating procedure for image analysis and experience, the analysts visually compared automated segmentation to manual segmentation and assessed whether the two segmentations achieved a good agreement (visually close to each other) or not. If there was a disagreement between the two, the analysts would score in three categories: automated segmentation performs better; manual segmentation performs better; not sure which one is better. The visual assessment was performed for basal, mid-ventricular and apical slices.

\subsection*{Exemplar clinical study}
We demonstrated the application of the method on an exemplar clinical study. Using automatically derived clinical measures, we investigated the association between cardiac function and obesity, similar to a previous research \cite{Rider2009}. We compared the ventricular volume and mass between two groups of subjects, the normal weight group (18.5 $\le$ body mass index (BMI) $<$ 25) and the obese group (BMI $\ge$ 30). Pathological cases with CVDs were excluded. The normal weight group and the obese group were matched for sex, age, height, diastolic blood pressure and systolic blood pressure using the nearest neighbour propensity score matching, implemented using the MatchIt package in R. After matching, each group consisted of 867 subjects. The clinical measures were then compared between the matched groups using two-sided t-tests.

\begin{figure*}[h!]
  \centering
  \includegraphics[width=16.5cm]{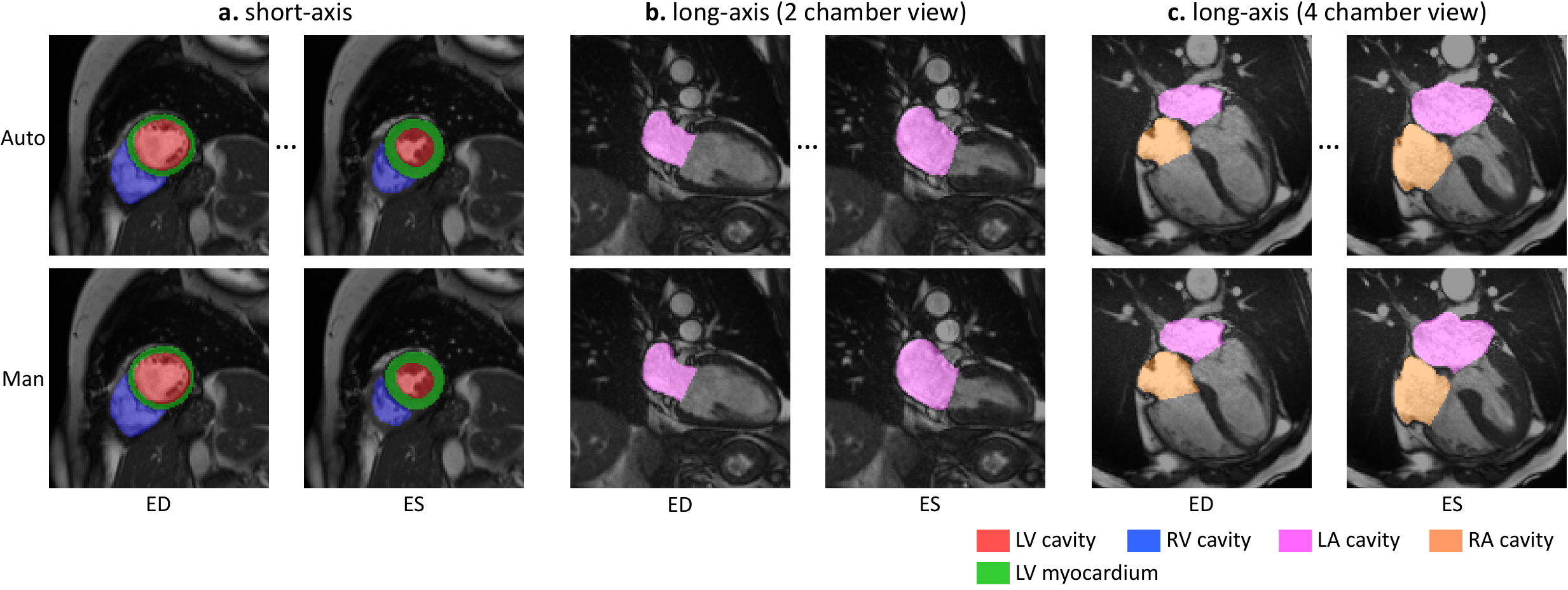}
  \caption{\textbf{Illustration of the segmentation results for short-axis and long-axis images.} The top row shows the automated segmentation, whereas the bottom row shows the manual segmentation. The automated method segments all the time frames. However, only end-diastolic (ED) and end-systolic (ES) frames are shown, as manual analysis only annotates ED and ES frames. The cardiac chambers are represented by different colours. \label{fig:seg}}
\end{figure*}

\section*{Results}
\subsection*{Short-axis image analysis}
Figure~\ref{fig:seg}a illustrates the predicted segmentation of the LV and RV on short-axis images. It shows that automated segmentation agrees well with manual segmentation by a clinical expert at both ED and ES time frames. Additional movie files demonstrate automated segmentation across a cardiac cycle [see Additional files 1-3].

Table~\ref{tab:dice_sa}(a) reports the Dice metric, mean contour distance and Hausdorff distance between automated and manual segmentations, evaluated on a test set of 600 subjects, which the network has never seen before. The table shows a mean Dice value of 0.94 for the LV cavity, 0.88 for the LV myocardium and 0.90 for the RV cavity, demonstrating a good agreement between automated and manual segmentations. The mean contour distance is 1.04 mm for the LV cavity, 1.14 mm for the LV myocardium and 1.78 mm for the RV cavity, all of which are smaller than the in-plane pixel spacing of 1.8 mm. The Hausdorff distance ranges from 3.16 mm to 7.25 mm for each class.

Of the 600 test subjects, 39 are with CVDs. These pathological cases were selected using the following criteria: cases with the International Classification of Diseases code, 10th Revision (ICD-10) of I21 (acute myocardial infarction), I22 (subsequent myocardial infarction), I23 (certain current complications following acute myocardial infarction), I25 (chronic ischaemic heart disease), I42 (cardiomyopathy), I50 (heart failure); cases where participants had self-reported heart attack. Table~\ref{tab:dice_sa}(b) reports the Dice and distance metrics on these pathological cases. It shows a consistent segmentation performance as on the full test set for the Dice metric and just slightly larger errors for the contour distance metrics.

\begin{table}[h!]
  \caption{\textbf{The Dice metric, mean contour distance (MCD) and Hausdorff distance (HD) between automated segmentation and manual segmentation for short-axis images.} The mean and standard deviation (in parenthesis) are reported. \label{tab:dice_sa}}
  \renewcommand{\arraystretch}{1.2}
  \centering
  \subfloat[The full test set (n = 600)]{
    \begin{tabular}{lccc}
    \hline 
    & Dice & MCD (mm) & HD (mm) \\
    \hline 
    LV cavity & 0.94 (0.04) & 1.04 (0.35) & 3.16 (0.98) \\
    LV myocardium & 0.88 (0.03) & 1.14 (0.40) & 3.92 (1.37) \\
    RV cavity & 0.90 (0.05) & 1.78 (0.70) & 7.25 (2.70) \\
    \hline
    \end{tabular}
  }\\
  \subfloat[Cases with CVDs (n = 39)]{
    \begin{tabular}{lccc}
    \hline 
    & Dice & MCD (mm) & HD (mm) \\
    \hline 
    LV cavity & 0.94 (0.04) & 1.19 (0.41) & 3.62 (1.14) \\
    LV myocardium & 0.87 (0.04) & 1.23 (0.40) & 4.28 (1.18) \\
    RV cavity & 0.90 (0.04) & 2.02 (0.88) & 8.19 (2.94) \\
    \hline
    \end{tabular}
  }\\
  \vspace{-1em}
  \begin{tabular}{@{\vspace{-1em}}p{\columnwidth}}
  CVD: cardiovascular diseases, LV: left ventricle, RV: right ventricle.
  \end{tabular}
\end{table}

For evaluating human performance, Table~\ref{tab:inter_sa} compares the Dice and distance metrics between automated segmentation and manual segmentation, as well as between segmentations by different human observers. It demonstrates that the computer-human difference is close to or even smaller than the human-human difference for all the metrics.

\begin{table*}[h!]
  \caption{\textbf{The Dice metric and contour distance metrics between automated segmentation and manual segmentation for short-axis images, as well between segmentations by different human observers.} The first column shows the difference between automated and manual segmentations on a test set of 600 subjects. The second to fourth columns show the inter-observer variability, which is evaluated on a randomly selected set of 50 subjects, each being analysed by three different human observers (O1, O2, O3) independently. The mean and standard deviation (in parenthesis) of the metrics are reported. \label{tab:inter_sa}}
  \renewcommand{\arraystretch}{1.2}
  \centering
  \subfloat[Dice metric]{
    \begin{tabular}{lcccc}
    \hline 
    & Auto vs Manual & O1 vs O2 & O2 vs O3 & O3 vs O1\\
    & (n = 600) & (n = 50) & (n = 50) & (n = 50) \\
    \hline
    LV cavity & 0.94 (0.04) & 0.94 (0.04) & 0.92 (0.04) & 0.93 (0.04) \\
    LV myocardium & 0.88 (0.03) & 0.88 (0.02) & 0.87 (0.03) & 0.88 (0.02) \\
    RV cavity & 0.90 (0.05) & 0.87 (0.06) & 0.88 (0.05) & 0.89 (0.05) \\
    \hline
    \end{tabular} 
  }
  
  \subfloat[Mean contour distance (mm)]{
    \begin{tabular}{lcccc}
    \hline
    & Auto vs Manual & O1 vs O2 & O2 vs O3 & O3 vs O1\\
    & (n = 600) & (n = 50) & (n = 50) & (n = 50) \\
    \hline 
    LV cavity & 1.04 (0.35) & 1.00 (0.25) & 1.30 (0.37) & 1.21 (0.48) \\
    LV myocardium & 1.14 (0.40) & 1.16 (0.34) & 1.19 (0.25) & 1.21 (0.36) \\
    RV cavity & 1.78 (0.70) & 2.00 (0.79) & 1.78 (0.45) & 1.87 (0.74) \\
    \hline
    \end{tabular}
  }
  
  \subfloat[Hausdorff distance (mm)]{
    \begin{tabular}{lcccc}
    \hline
    & Auto vs Manual & O1 vs O2 & O2 vs O3 & O3 vs O1\\
    & (n = 600) & (n = 50) & (n = 50) & (n = 50) \\
    \hline 
    LV cavity & 3.16 (0.98) & 2.84 (0.70) & 3.31 (0.90) & 3.25 (0.96)  \\
    LV myocardium & 3.92 (1.37) & 3.70 (1.16) & 3.82 (1.07) & 3.76 (1.21) \\
    RV cavity & 7.25 (2.70) & 7.56 (2.51) & 7.35 (2.19) & 7.14 (2.20) \\
    \hline
    \end{tabular}
  }
\end{table*}

\begin{table*}[h!]
  \caption{\textbf{Qualitative visual assessment of automated segmentation.} Two experienced image analysts visually compared automated segmentation to manual segmentation for 250 test subjects and assessed whether the two segmentations achieved a good agreement (visually close to each other) or not. If there was a disagreement between the two, the analysts would score in three categories: automated segmentation performs better; manual segmentation performs better; not sure which one is better. The visual assessment was performed for basal, mid-ventricular and apical slices. The percentage of each score catetory is reported. \label{tab:visual}}
  \renewcommand{\arraystretch}{1.2}
  \centering
  \begin{tabular}{llcccc}
  \hline 
  & & \multirow{2}{*}{Agreement (\%)} & \multicolumn{3}{c}{Disagreement (\%)} \\
  \cline{4-6}
  & & & Auto. better & Man. better & Not sure \\
  \hline 
  Analyst 1 & Basal & 40.0 & 26.2 & 20.6 & 13.2 \\ 
  & Mid-ventricular & 84.8 & 12.2 & 2.4 & 0.6 \\
  & Apical & 44.0 & 29.0 & 22.0 & 5.0 \\
  \hline 
  Analyst 2 & Basal & 33.0 & 27.4 & 17.4 & 22.2 \\ 
  & Mid-ventricular & 91.6 & 6.6 & 1.8 & 0.0 \\
  & Apical & 80.8 & 8.8 & 9.6 & 0.8 \\
  \hline
  \end{tabular}
\end{table*}

\begin{table*}[h!]
  \caption{\textbf{The difference in clinical measures between automated segmentation and manual segmentation, as well between measurements by different human observers.} The first column shows the difference between automated and manual segmentations on a test set of 600 subjects. The second to fourth columns show the inter-observer variability, which is evaluated on a randomly selected set of 50 subjects, each being analysed by three different human observers (O1, O2, O3) independently. The mean and standard deviation (in parenthesis) of the absolute difference and relative difference are reported. \label{tab:clin}}
  \renewcommand{\arraystretch}{1.2}
  \centering
  \subfloat[Absolute difference]{
    \begin{tabular}{lcccc}
    \hline 
    & Auto vs Manual & O1 vs O2 & O2 vs O3 & O3 vs O1\\
    & (n = 600) & (n = 50) & (n = 50) & (n = 50) \\
    \hline 
    LVEDV (mL) & 6.1 (5.3) & 6.1 (4.4)  & 8.8 (4.8) & 4.8 (3.1) \\
    LVESV (mL) & 5.3 (4.9) & 4.1 (4.2)  & 6.7 (4.2) & 7.1 (3.8) \\
    LVM (gram) & 6.9 (5.5) & 4.2 (3.2)  & 6.6 (4.9) & 6.5 (4.8) \\
    RVEDV (mL) & 8.5 (7.1) & 11.1 (7.2) & 6.2 (4.6) & 8.7 (5.8)  \\
    RVESV (mL) & 7.2 (6.8) & 15.6 (7.8) & 6.6 (5.5) & 11.7 (6.9) \\
    \hline
    \end{tabular}
  }\\
  \subfloat[Relative difference]{
    \begin{tabular}{lcccc}
    \hline 
    & Auto vs Manual & O1 vs O2 & O2 vs O3 & O3 vs O1\\
    & (n = 600) & (n = 50) & (n = 50) & (n = 50) \\
    \hline 
    LVEDV (\%) & 4.1 (3.5) & 4.2 (3.1) & 6.3 (3.3) & 3.4 (2.2) \\
    LVESV (\%) & 9.5 (9.5) & 6.8 (7.5) & 12.5 (8.5) & 11.7 (5.1) \\
    LVM (\%) & 8.3 (7.6) & 4.4 (3.3) & 6.0 (3.7) & 6.7 (4.6) \\
    RVEDV (\%) & 5.6 (4.6) & 8.0 (5.0) & 4.2 (3.1) & 5.7 (3.6) \\
    RVESV (\%) & 11.8 (12.2) & 30.6 (15.5) & 10.9 (8.3) & 16.9 (9.2) \\
    \hline
    \end{tabular} 
  }
\end{table*}

As an additional qualitative assessment, two image analysts visually compared automated segmentation to manual segmentation for 250 test subjects. Table~\ref{tab:visual} shows that for mid-ventricular slices, automated segmentation agrees well with manual segmentation for respectively 84.8\% and 91.6\% of the cases by visual inspection of the two analysts. For basal slices where the ventricular contours are more complex and thus more difficult to segment, the percentage of agreement is lower. For example, Analyst 1 scored that automated segmentation agrees well with manual segmentation for only 40.0\% of the cases. When discrepancy occurs, however, automated segmentation performs similarly to manual segmentation. Analyst 1 scored that automated segmentation performs better for 26.2\% of the cases, whereas manual segmentation performs better for 20.6\% of the cases.

Next, we evaluate the accuracy of clinical measures for the LVEDV, LVESV, LVM, RVEDV and RVESV. Table~\ref{tab:clin} reports the mean absolute difference and relative difference between automated and manual measurements and between measurements by different expert observers. It shows that for the clinical measures, the computer-human difference is on par with the human-human difference.

\begin{figure*}[h!]
  \centering
  \includegraphics[width=16.5cm]{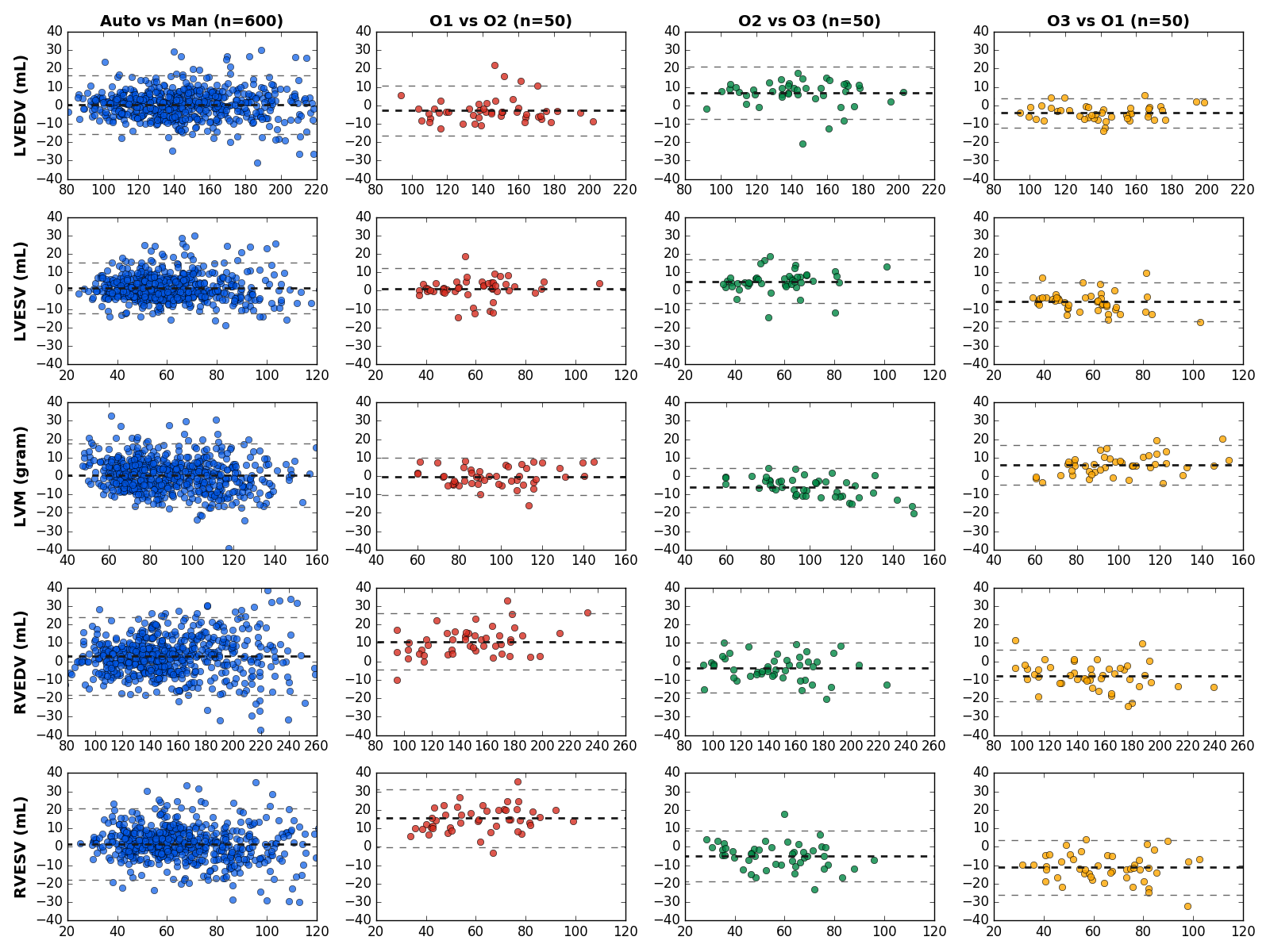}
  \caption{\textbf{Bland-Altman plots of clinical measures between automated measurement and manual measurement, as well between measurements by different human observers.} The first column shows the agreement between automated and manual measurements on a test set of 600 subjects. The second to fourth columns show the inter-observer variability evaluated on the randomly selected set of 50 subjects. In each Bland-Altman plot, the x-axis denotes the average of two measurements and the y-axis denotes the difference between them. The dark dashed line denotes the mean difference (bias) and the two light dashed lines denote $\pm$1.96 standard deviations from the mean. \label{fig:bland_altman}}
\end{figure*}

Figure~\ref{fig:bland_altman} shows the Bland-Altman plots of the clinical measures. The Bland-Altman plot is commonly used for analysing agreement and bias between two measurements. The first column of the figure compares automated measurements to manual measurements on 600 test subjects. These subjects were annotated by a group of eight observers and each subject was annotated only once by one observer. The first column shows that the mean difference is centred close to zero, which suggests that the automated measurement is almost unbiased relative to the group of observers. Also, there is no evidence of bias over hearts of difference sizes or volumes. By contrast, the bias between different pairs of human observers (second to fourth columns) is often larger than that, especially for RVEDV and RVESV. This indicates that individual observers may be biased. As the automated method is trained with annotations from multiple observers, it learns a consensus estimate across the group of observers and thus it may be less susceptible to biases.

\subsection*{Long-axis image analysis}
We further demonstrate the performance of the method on long-axis CMR images, which are commonly used for assessing the cardiac chambers from a different angle. Figures~\ref{fig:seg}b and \ref{fig:seg}c illustrate the segmentations of the LA and RA for the long-axis 2Ch and 4Ch images respectively. Additional movie files demonstrate automated segmentation across a cardiac cycle [see Additional files 4-5].

We evaluate the Dice metric and the contour distances on a test set of 600 subjects, as reported in Table~\ref{tab:dice_la}. The mean Dice metric is 0.93 for the LA (2Ch), 0.95 for the LA (4Ch), 0.96 for the RA (4Ch), whereas the mean contour distance is smaller than the in-plane pixel spacing of 1.8 mm, demonstrating a good segmentation accuracy on long-axis images. Table~\ref{tab:inter_la} demonstrates that for long-axis images, the computer-human difference is also on par with or smaller than the human-human difference.

\begin{table}[h!]
  \caption{\textbf{The Dice metric, mean contour distance (MCD) and Hausdorff distance (HD) between automated segmentation and manual segmentation for long-axis images.} The mean and standard deviation (in parenthesis) are reported on a test set of 600 subjects. \label{tab:dice_la}}
  \renewcommand{\arraystretch}{1.2}
  \centering
  \begin{tabular}{lccc}
  \hline 
  & Dice & MCD (mm) & HD (mm) \\
  \hline 
  LA cavity (2Ch) & 0.93 (0.05) & 1.46 (1.06) & 5.76 (5.85) \\
  LA cavity (4Ch) & 0.95 (0.02) & 1.04 (0.38) & 4.03 (2.26) \\
  RA cavity (4Ch) & 0.96 (0.02) & 0.99 (0.43) & 3.89 (2.39) \\
  \hline
  \end{tabular} 
  \begin{tabular}{@{\vspace{-1em}}p{\columnwidth}}
  LA: left atrium, RA: right atrium.
  \end{tabular}
\end{table}

\begin{table*}[h!]
  \caption{\textbf{The Dice metric and contour distance metrics between automated segmentation and manual segmentation for long-axis images, as well between segmentations by different human observers.} The first column shows the difference between automated and manual segmentations on a test set of 600 subjects. The second to fourth columns show the inter-observer variability, which is evaluated on a randomly selected set of 50 subjects, each being analysed by three different human observers (O1, O2, O3) independently. The mean and standard deviation (in parenthesis) of the metrics are reported. \label{tab:inter_la}}
  \renewcommand{\arraystretch}{1.2}
  \centering
  \subfloat[Dice metric]{
    \begin{tabular}{lcccc}
    \hline 
    & Auto vs Manual & O1 vs O2 & O2 vs O3 & O3 vs O1\\
    & (n = 600) & (n = 50) & (n = 50) & (n = 50) \\
    \hline
    LA cavity (2Ch) & 0.93 (0.05) & 0.92 (0.02) & 0.90 (0.04) & 0.90 (0.04) \\
    LA cavity (4Ch) & 0.95 (0.02) & 0.95 (0.03) & 0.94 (0.02) & 0.94 (0.03) \\
    RA cavity (4Ch) & 0.96 (0.02) & 0.95 (0.02) & 0.95 (0.02) & 0.95 (0.02) \\
    \hline
    \end{tabular} 
  }
  
  \subfloat[Mean contour distance (mm)]{
    \begin{tabular}{lcccc}
    \hline
    & Auto vs Manual & O1 vs O2 & O2 vs O3 & O3 vs O1\\
    & (n = 600) & (n = 50) & (n = 50) & (n = 50) \\
    \hline 
    LA cavity (2Ch) & 1.46 (1.06) & 1.57 (0.39) & 1.94 (0.68) & 1.95 (0.57) \\
    LA cavity (4Ch) & 1.04 (0.38) & 1.08 (0.40) & 1.21 (0.33) & 1.23 (0.35) \\
    RA cavity (4Ch) & 0.99 (0.43) & 1.13 (0.35) & 1.22 (0.37) & 1.16 (0.37) \\
    \hline
    \end{tabular}
  }
  
  \subfloat[Hausdorff distance (mm)]{
    \begin{tabular}{lcccc}
    \hline
    & Auto vs Manual & O1 vs O2 & O2 vs O3 & O3 vs O1\\
    & (n = 600) & (n = 50) & (n = 50) & (n = 50) \\
    \hline 
    LA cavity (2Ch) & 5.76 (5.85) & 5.66 (1.97) & 7.16 (3.12) & 6.78 (2.53) \\
    LA cavity (4Ch) & 4.03 (2.26) & 3.89 (1.85) & 4.29 (1.97) & 4.06 (1.44) \\
    RA cavity (4Ch) & 3.89 (2.39) & 4.31 (2.20) & 4.20 (2.16) & 4.08 (2.06) \\
    \hline
    \end{tabular}
  }
\end{table*}

\begin{table}[h!]
  \caption{\textbf{An exemplar study of cardiac function on large-scale datasets using automatically derived clinical measures.}  It compares the normal weight group (18.5 $\le$ BMI $<$ 25) to the obese group (BMI $\ge$ 30). The mean and standard deviation (in parenthesis) are reported. \label{tab:asso}}
  \renewcommand{\arraystretch}{1.2}
  \centering
    \begin{tabular}{lccc}
    \hline 
    & Normal & Obese & \multirow{2}{*}{p-value} \\
    & (n = 867) & (n = 867) & \\
    \hline 
    LVEDV (mL) & 143 (31) & 158 (34) & $<$0.001 \\
    LVESV (mL) & 60 (19) & 67 (20) & $<$0.001 \\
    LVM (gram) & 85 (20) & 103 (26) & $<$0.001 \\
    RVEDV (mL) & 152 (36) & 167 (38) & $<$0.001 \\
    RVESV (mL) & 67 (20) & 75 (22) & $<$0.001 \\
    \hline
    \end{tabular}
    \begin{tabular}{@{\vspace{-1em}}p{\columnwidth}}
    BMI: body mass index, LVEDV: left ventricular end-diastolic volume, LVESV: left ventricular end-systolic volume, LVM: left ventricular mass, RVEDV: right ventricular end-diastolic volume, RVESV: right ventricular end-systolic volume.
    \end{tabular}
\end{table}

\subsection*{Exemplar clinical study}
The proposed automated method enables us to perform clinical studies on large-scale datasets. 
Table~\ref{tab:asso} compares the ventricular volume and mass, which are derived from automated segmentation, between two groups of subjects, the normal weight group and the obese group. The table shows that obesity is associated with increased ventricular volume and mass with statistical significance. This is consistent with a previous finding in \cite{Rider2009}, which was performed on a dataset of 54 subjects with manual segmentation. Now we can confirm the finding with automated analysis on a much larger dataset with 1,734 subjects.

\section*{Discussion}
By training and evaluating on a large-scale annotated dataset, we demonstrate that the proposed method matches human expert performance on CMR image segmentation accuracy and clinical measurement accuracy. In terms of speed, it can analyse the short-axis and long-axis images for one subject in a few seconds. The method is fast and scalable, overcoming limitations associated with current clinical CMR image analysis routine, which is manual, time-consuming and prone to subjective errors. The method has a great potential for improving work efficiency and assisting clinicians in diagnosis and performing large-scale clinical research.

\subsection*{Residual networks}
We also experimented with a deeper network by replacing the convolutional layers from scale 3 to 5 in Table~\ref{tab:net_config} with residual blocks as described in \cite{He2016} and constructed a residual network which has 33 convolutional layers. In experiments, we found the residual network achieves a similar performance as the VGG-16 network. Thus, we only reported the results from the VGG-16 network in the paper.

\subsection*{Other clinical measures}
The LV and RV volumes are directly calculated from the image segmentations. There are also some other clinical measures for assessing cardiac function, which are derived from the LV and RV volumes, including the LV stroke volume (LVSV), LV ejection fraction (LVEF), LV cardiac output (LVCO), RV stroke volume (RVSV), RV ejection fraction (RVEF) and RV cardiac output (RVCO). Table~\ref{tab:derived_clin} reports the difference between automated and manual measurements and between measurements by different expert observers on these measures. It shows that for these derived clinical measures, the computer-human difference is also comparable to the human-human difference.

\begin{table*}[h!]
  \caption{\textbf{The difference in derived clinical measures between automated segmentation and manual segmentation, as well between measurements by different human observers.} The first column shows the difference between automated and manual segmentations on a test set of 600 subjects. The second to fourth columns show the inter-observer variability, which is evaluated on a randomly selected set of 50 subjects, each being analysed by three different human observers (O1, O2, O3) independently. The mean and standard deviation (in parenthesis) of the absolute difference and relative difference are reported. \label{tab:derived_clin}}
  \renewcommand{\arraystretch}{1.2}
  \centering
  \subfloat[Absolute difference]{
    \begin{tabular}{lcccc}
    \hline 
    & Auto vs Manual & O1 vs O2 & O2 vs O3 & O3 vs O1\\
    & (n = 600) & (n = 50) & (n = 50) & (n = 50) \\
    \hline 
    LVSV (mL) & 6.1 (5.6) & 6.6 (4.1) & 5.6 (4.1) & 4.2 (3.2) \\
    LVEF (\%) & 3.2 (2.9) & 3.1 (2.1) & 3.0 (2.4) & 3.8 (1.8) \\
    LVCO (L/min) & 0.4 (0.3) & 0.4 (0.2) & 0.3 (0.2) & 0.3 (0.2) \\
    RVSV (mL) & 8.1 (6.8) & 7.1 (5.5) & 5.3 (4.2) & 5.4 (4.8) \\
    RVEF (\%) & 4.3 (3.6) & 7.8 (4.4) & 3.7 (2.7) & 5.7 (3.9) \\
    RVCO (L/min) & 0.5 (0.4) & 0.4 (0.3) & 0.3 (0.2) & 0.3 (0.3) \\
    \hline
    \end{tabular}
  }\\
  \subfloat[Relative difference]{
    \begin{tabular}{lcccc}
    \hline 
    & Auto vs Manual & O1 vs O2 & O2 vs O3 & O3 vs O1\\
    & (n = 600) & (n = 50) & (n = 50) & (n = 50) \\
    \hline 
    LVSV (\%) & 7.0 (5.8) & 7.4 (4.1) & 6.5 (4.8) & 4.8 (3.3) \\
    LVEF (\%) & 5.4 (4.8) & 5.1 (3.7) & 4.9 (3.8) & 6.6 (3.2) \\
    LVCO (\%) & 7.0 (5.8) & 7.4 (4.1) & 6.5 (4.8) & 4.8 (3.3) \\
    RVSV (\%) & 9.6 (8.3) & 8.1 (6.9) & 6.1 (4.4) & 7.1 (8.5) \\
    RVEF (\%) & 7.5 (6.2) & 12.3 (6.6) & 6.5 (5.0) & 10.7 (7.9) \\
    RVCO (\%) & 9.6 (8.3) & 8.1 (6.9) & 6.1 (4.4) & 7.1 (8.5) \\
    \hline
    \end{tabular} 
  }\\
  \vspace{-1em}
  \begin{tabular}{p{\linewidth}}
  LVSV: left ventricular stroke volume, LVEF: left ventricular ejection fraction, LVCO: left ventricular cardiac output, RVSV: right ventricular stroke volume, RVEF: right ventricular ejection fraction, RVCO: right ventricular cardiac output.
  \end{tabular}
\end{table*}

\subsection*{Limitations}
A major limitation of our work is that the neural network was trained on a single dataset, the UK Biobank dataset, which is a relatively homogeneous dataset. The majority of the data are healthy subjects in middle and later life and only a small proportion are with self-reported cardiovascular diseases \cite{Fry2017}. Although we have demonstrated that the method works well on a subset of pathological cases in Table~\ref{tab:dice_sa}(b), in the clinical environment, there can be a variety of pathological patterns, which are not currently represented in the UK Biobank cohort.

In addition, the UK Biobank dataset was acquired using a standard imaging protocol and the same scanner model \cite{Petersen2016}. This guarantees that the derived image phenotypes are consistent across the UK Biobank study, without being biased by the imaging protocol or the scanner model. However, this also means that the neural network that we have learnt is adapted to the image patterns in the UK Biobank dataset and might not generalise well to other vendor or sequence datasets. We explored how the network works on two additional datasets, the MICCAI 2009 Left Ventricle Segmentation Challenge (LVSC 2009) dataset \cite{Radau2009} and the MICCAI 2017 Automated Cardiac Diagnosis Challenge (ACDC 2017) dataset \cite{Bernard2018}. These two datasets were acquired using different scanners or different protocols \cite{LVSC2009, ACDC2017} from the UK Biobank dataset. In addition, most of the LVSC 2009 and ACDC 2017 data are pathological cases.

\begin{figure*}[h!]
  \centering
  \includegraphics[width=13.5cm]{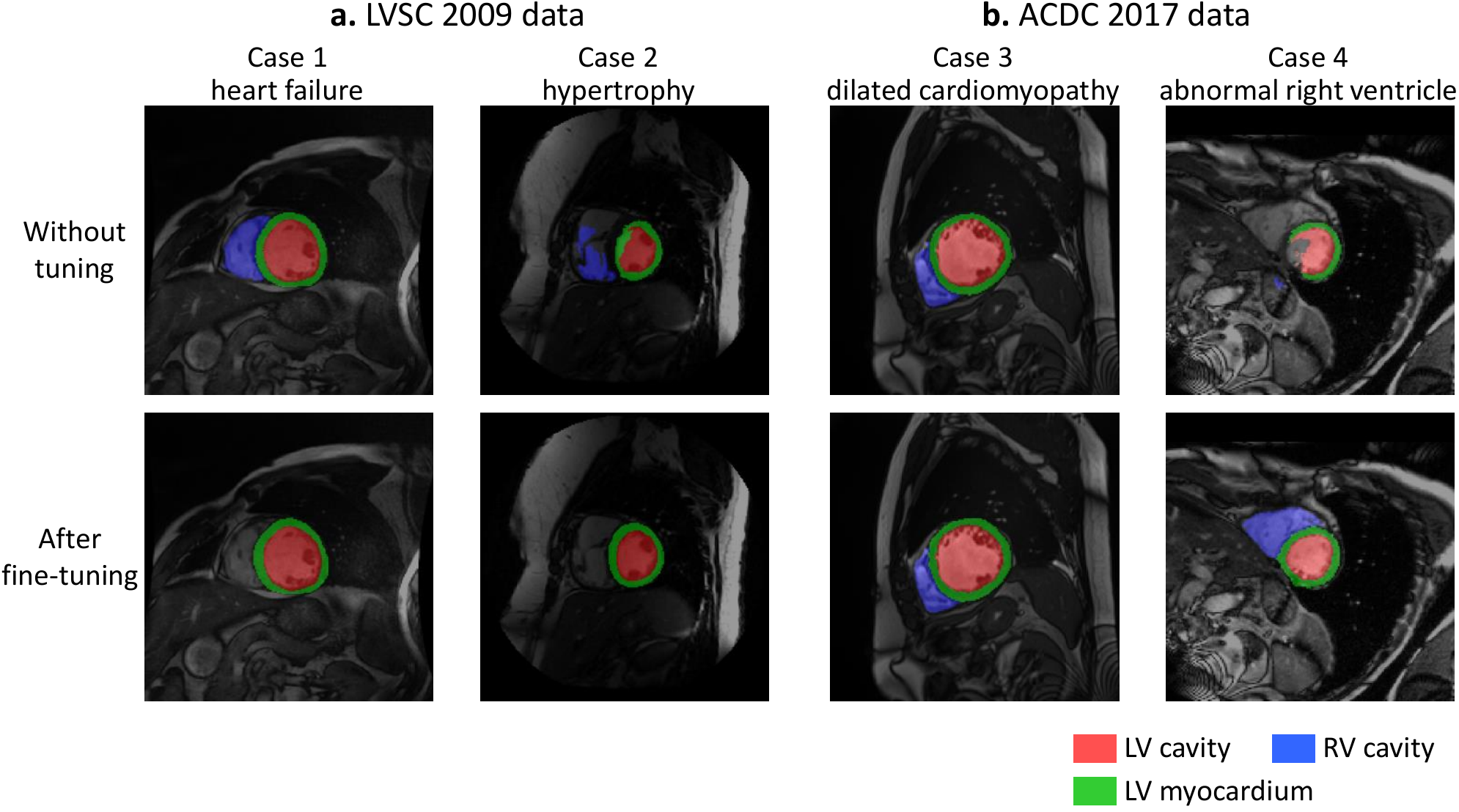}
  \caption{\textbf{Segmentation results on other datasets.} The first two cases come from the LVSC 2009 dataset, whereas the last two cases come from the ACDC 2017 dataset. The four cases are respectively of heart failure, LV hypertrophy, dilated cardiomyopathy and abnormal right ventricle. The top row shows the segmentation results by directly applying the UK Biobank-trained network to the LVSC and ACDC data. The bottom row shows the segmentation results after fine-tuning the network to the new data. \label{fig:general}}
\end{figure*}

Figure~\ref{fig:general} shows the segmentation results of four exemplar cases, two from the LVSC 2009 dataset and two from the ACDC 2017 dataset. The four cases are respectively of heart failure, LV hypertrophy, dilated cardiomyopathy and abnormal right ventricle. The top row shows the segmentation results by directly applying the UK Biobank-trained network to the LVSC and ACDC data. It shows that without any tuning, the network performs well for Cases 1 and 3, but fails for Cases 2 and 4. This is probably because the image patterns or intensity distributions in Cases 2 and 4 are not covered by UK Biobank.

Then, we performed fine-tuning for the network by training it for another 10,000 iterations on the new datasets, which took about 2 hour. For LVSC 2009, we fine-tuned using the challenge training set (15 subjects) and evaluated the performance on the challenge validation set (15 subjects).  The LVSC 2009 training set only annotates the LV cavity and myocardium. As a result, during fine-tuning, we only trained the network to segment the LV and ignored the RV. For ACDC 2017, we randomly split the challenge training set (100 subjects) into 80 subjects for fine-tuning and 20 subjects for evaluation. The bottom row of Figure~\ref{fig:general} shows the segmentation results on LVSC or ACDC data after fine-tuning. It shows that the segmentation performance is substantially improved for Cases 2 and 4 after the network has adjusted its parameters to adapt to the new data. Table~\ref{tab:general} reports the Dice overlap metrics before and after fine-tuning. On both LVSC\footnote{We evaluated the Dice metric between automated and manual segmentions in 3D. Previous studies on LVSC may report the Dice metric for good contours only (with distance error less than 5mm) \cite{Ngo2013}.} and ACDC datasets, the Dice metrics are substantially improved after fine-tuning.

\begin{table*}[h!]
  \caption{\textbf{Dice overlap metrics for segmentations on LVSC 2009 and ACDC 2017 datasets.} The performances using the UK Biobank-trained network without fine-tuning and after fine-tuning are compared. The mean and standard deviation (in parenthesis) are reported. \label{tab:general}}
  \renewcommand{\arraystretch}{1.2}
  \centering
  \begin{tabular}{lcccc}
  \hline 
  & \multicolumn{2}{c}{LVSC 2009} & \multicolumn{2}{c}{ACDC 2017} \\
  & \multicolumn{2}{c}{validation set (n = 15)} & \multicolumn{2}{c}{training set split (n = 20)} \\
  \cline{2-5}
  & w.o. fine-tune & w. fine-tune & w.o. fine-tune & w. fine-tune \\
  \hline
  LV cavity & 0.72 (0.22) & 0.90 (0.08) & 0.74 (0.29) & 0.94 (0.04) \\
  LV myocardium & 0.56 (0.18) & 0.81 (0.05) & 0.65 (0.24) & 0.88 (0.05) \\
  RV cavity & - & - & 0.60 (0.35) & 0.88 (0.08) \\
  \hline
  \end{tabular} 
\end{table*}

Although the network works well after fine-tuning, this still means each time when we have some new data that are acquired using a different protocol or from a different scanner model, we might need to label some of the new data for fine-tuning the network parameters. It would be interesting to explore whether we could create a large-scale heterogeneous dataset for training and evaluation, which covers typical CMR imaging protocols and scanner types, or to develop novel machine learning techniques that are more generalisable, which is an important research topic on its own \cite{Marcus2018}.

\subsection*{Future directions}
Future research will explore developing more generalisable methods for analysing a wider range of CMR images, such as multi-site images acquired from different machines and using different imaging protocols, and integrating automated segmentation results into diagnostic reports. The current method trains networks for short-axis images and long-axis images separately. It would be interesting to combine the two views for image analysis, which can provide complementary information about the anatomy of the heart. Finally, we believe that a benchmark platform based on this annotated dataset is needed, which would benefit the whole community and greatly advance the development of CMR image analysis algorithms.

\section*{Conclusions}
We have proposed an automated method using deep FCN for short-axis and long-axis CMR image analysis. It has demonstrated a human-level performance on the UK Biobank dataset. We anticipate this to be a starting point for automated CMR analysis, facilitated by machine learning.


\begin{backmatter}

\section*{Abbreviations}
BMI: body mass index; bSSFP: balanced steady-state free precession; CMR: cardiovascular magnetic resonance; CT: computed tomography; CVD: cardiovascular disease; ED: end-diastole; ES: end-systole; FCN: fully convolutional network; GPU: graphics processing unit; HD: Hausdorff distance; ICD-10: International Classification of Diseases code, 10th Revision; LA: left atrium; LV: left ventricle; LVCO: left ventricular cardiac output; LVEDV: left ventricular end-diastolic volume; LVEF: left ventricular ejection fraction; LVESV: left ventricular end-systolic volume; LVM: left ventricular mass; LVSV: left ventricular stroke volume; MCD: mean contour distance; RA: right atrium; RV: right ventricle; RVCO: right ventricular cardiac output; RVEDV: right ventricular end-diastolic volume; RVEF: right ventricular ejection fraction; RVESV: right ventricular end-systolic volume; RVSV: right ventricular stroke volume; 2Ch: 2-chamber view; 4Ch: 4-chamber view.

\section*{Ethics approval and consent to participate}
UK Biobank has approval from the North West Research Ethics Committee (REC reference: 11/NW/0382). 

\section*{Consent for publication}
Not applicable.

\section*{Availability of data and material}
The imaging data and manual annotations were provided by the UK Biobank Resource under Application Number 2946. Researchers can apply to use the UK Biobank data resource for health-related research in the public interest \cite{UKBiobankReg2017}. The image analysis source code is available at \url{https://github.com/baiwenjia/ukbb_cardiac}. The code is used for data format conversion, pre-processing, segmentation network training, testing and clinical measure calculation.

\section*{Competing interests}
S.E.P. receives consultancy fees from Circle Cardiovascular Imaging Inc., Calgary, Alberta, Canada.

\section*{Funding}
This work is supported by the SmartHeart EPSRC Programme Grant (EP/P001009/1). G.T. is supported by a Marie Sk\l{}odowska Curie European Fellowship. A.L. and S.E.P. acknowledge support from the NIHR Barts Biomedical Research Centre and from the MRC for the MRC eMedLab Medical Bioinformatics infrastructure (MR/L016311/1), which enables data access. N.A. is supported by a Wellcome Trust Research Training Fellowship (203553/Z/Z). S.N. and S.K.P. acknowledge support from the NIHR Oxford Biomedical Research Centre and the Oxford BHF Centre of Research Excellence. S.E.P., S.K.P. and S.N. acknowledge the British Heart Foundation (BHF) for funding the manual analysis to create a cardiovascular magnetic resonance imaging reference standard for the UK Biobank imaging resource in 5000 CMR scans (PG/14/89/31194). H.S. is supported by a Research Fellowship from the Uehara Memorial Foundation. P.M.M. gratefully acknowledges support from the Edmond J. Safra Foundation and Lily Safra, the Imperial College Healthcare Trust Biomedical Research Centre, the EPSRC Centre for Mathematics in Precision Healthcare and the MRC.

\section*{Authors' contributions}
W.B., B.G. and D.R. conceived and designed the study; M.S., G.T., O.O., M.R., and G.V. provided advice and support on computing method aspects; S.N., S.E.P., S.K.P. provided the design of a large data resource to be used for training and testing of artificial intelligence approaches; A.M.L., N.A., S.E.P., S.K.P. and S.N. provided advice and support on clinical aspects; N.A., E.L., M.M.S., F.Z., K.F., J.M.P., V.C. and Y.J.K. performed manual image annotation under the senior supervision of S.E.P., S.K.P. and S.N.; E.L. and K.F. performed qualitative visual assessment of automated segmentation; A.M.L. and V.C. curated the annotation database; H.S. and P.M.M. provided advice and support in the initial stage of model development; W.B., B.K. and A.M.L. performed data pre-processing; W.B. designed the method, performed data analysis and wrote the manuscript. All authors read and approved the manuscript.

\section*{Acknowledgements}
This research has been conducted mainly using the UK Biobank Resource under Application Number 2946. The initial stage of the research was conducted using the UK Biobank Resource under Application Number 18545. The authors wish to thank all UK Biobank participants and staff.


\bibliographystyle{bmc-mathphys} 
\bibliography{refs}      




\section*{Additional Files}
\subsection*{Additional file 1 - Movie demonstrating short-axis image segmentation (mid-ventricular slice)}

\subsection*{Additional file 2 - Movie demonstrating short-axis image segmentation (basal slice)}

\subsection*{Additional file 3 - Movie demonstrating short-axis image segmentation (apical slice)}

\subsection*{Additional file 4 - Movie demonstrating long-axis image segmentation (2 chamber view)}

\subsection*{Additional file 5 - Movie demonstrating long-axis image segmentation (4 chamber view)}

\subsection*{Additional file 6 - Image demonstrating visual assessment and comparison between automated segmentation and manual segmentation}

\end{backmatter}

\end{document}